\documentclass{article}

\usepackage{arydshln}
\usepackage{caption}
\usepackage{longtable}
\usepackage{microtype}
\usepackage{graphicx}
\usepackage{subfigure}
\usepackage{booktabs} % for professional tables

\usepackage{hyperref}

\usepackage[accepted]{icml2024}

\usepackage{amsmath}
\usepackage{amssymb}
\usepackage{mathtools}
\usepackage{amsthm}

\usepackage{algorithm}
\usepackage{algorithmic}
\usepackage{multirow}
\usepackage{adjustbox}
\usepackage{placeins}
\usepackage{pifont}
\usepackage[capitalize,noabbrev]{cleveref}

\theoremstyle{plain}

\theoremstyle{definition}

\theoremstyle{remark}

\usepackage[textsize=tiny]{todonotes}

\icmltitlerunning{}

\begin{document}

% T-RevSNN: 

\twocolumn[
\icmltitle{High-Performance Temporal Reversible Spiking Neural Networks with  $\mathcal{O}(L)$ Training Memory and $\mathcal{O}(1)$ Inference Cost}



\icmlsetsymbol{equal}{$*$}
\icmlsetsymbol{corresponding author}{$\dag$}

\begin{icmlauthorlist}
\icmlauthor{Jiakui Hu$^{*}$}{pku}
\icmlauthor{Yao Man$^{*}$}{ia}
\icmlauthor{Xuerui Qiu}{iaf}
\icmlauthor{Yuhong Chou}{polyu}
\icmlauthor{Yuxuan Cai}{01ai}
\icmlauthor{Ning Qiao}{syns}
\icmlauthor{Yonghong Tian}{pku,pc}
\icmlauthor{Bo Xu}{ia}
\icmlauthor{Guoqi Li$^{\dag}$}{ia,BICLab}
\end{icmlauthorlist}

\icmlaffiliation{pku}{Peking University, Beijing, China}
\icmlaffiliation{ia}{Institute of Automation, Chinese Academy of Sciences, Beijing, China}
\icmlaffiliation{pc}{Peng Cheng Laboratory, Shenzhen, Guangzhou, China}
\icmlaffiliation{BICLab}{Key Laboratory of Brain Cognition and Brain-inspired Intelligence Technology, Beijing, China}
\icmlaffiliation{iaf}{School of Future Technology, University of Chinese Academy of Sciences}
\icmlaffiliation{syns}{SynSense AG Corporation, Zurich, Switzerland}
\icmlaffiliation{polyu}{Department of Computing, The Hong Kong Polytechnic University, HongKong, China}
\icmlaffiliation{01ai}{01.AI, Beijing, China}

\icmlcorrespondingauthor{Guoqi Li}{guoqi.li@ia.ac.cn}

\icmlkeywords{Machine Learning, ICML}

\vskip 0.1in
]



\printAffiliationsAndNotice{\icmlEqualContribution} % otherwise use the standard text.

\begin{abstract}
Multi-timestep simulation of brain-inspired Spiking Neural Networks (SNNs) boost memory requirements during training and increase inference energy cost. Current training methods cannot simultaneously solve both training and inference dilemmas. This work proposes a novel Temporal Reversible architecture for SNNs (T-RevSNN) to jointly address the training and inference challenges by altering the forward propagation of SNNs. We turn off the temporal dynamics of most spiking neurons and design multi-level temporal reversible interactions at temporal turn-on spiking neurons, resulting in a $\mathcal{O}(L)$ training memory. Combined with the temporal reversible nature, we redesign the input encoding and network organization of SNNs to achieve $\mathcal{O}(1)$ inference energy cost. Then, we finely adjust the internal units and residual connections of the basic SNN block to ensure the effectiveness of sparse temporal information interaction. T-RevSNN achieves excellent accuracy on ImageNet, while the memory efficiency, training time acceleration, and inference energy efficiency can be significantly improved by $8.6 \times$, $2.0 \times$, and $1.6 \times$, respectively. This work is expected to break the technical bottleneck of significantly increasing memory cost and training time for large-scale SNNs while maintaining high performance and low inference energy cost. Source code and models are available at: \url{https://github.com/BICLab/T-RevSNN}.
\end{abstract}

\section{Introduction}
Spiking Neural Networks (SNNs) are promised to be a low-power alternative to Artificial Neural Networks (ANNs), thanks to their brain-inspired neuronal dynamics and spike-based communication \cite{Nature_2,schuman2022opportunities}. Spiking neurons fuse spatio-temporal information through intricate dynamics and fire binary spikes as outputs when their membrane potential crosses the threshold \cite{maass1997networks}. Spike-based communication enables the event-driven computational paradigm of SNNs and has extremely high power efficiency in neuromorphic chips such as TrueNorth \cite{2014TrueNorth}, Loihi \cite{davies2018loihi}, Tianjic \cite{Nature_1}, and Speck \cite{Speck}. 

By discretizing the nonlinear differential equation of spiking neurons into iterative versions \cite{wu2018spatio,neftci2019surrogate}, SNNs can be trained directly using BackPropagation Through Time (BPTT) and surrogate gradient \cite{wu2018spatio,neftci2019surrogate}. Multi-timestep simulation requires a memory of $\mathcal{O}(L \times T)$ for SNN training, where $L$ is the layer and $T$ is the timestep. In particular, in static vision tasks, such as image classification, it is common practice \cite{wu2019direct,deng2020rethinking,9747906,yao_attention_Pami,9556508} to exploit the dynamics of SNNs by repeating the inputs to achieve high performance. In this case, the inference energy is $\mathcal{O}(T)$.

As an example, training spiking ResNet-19 with 10 timestep takes about $20\times$ more memory than ResNet-19 \cite{fang2023spikingjelly}. Numerous efforts have been made to solve the memory dilemma. The mainstream idea is to decouple the training of SNNs from the timestep, e.g., modify the way the BPTT in the temporal \cite{xiao2022online,meng2023towards,10242251}, training with $T=1$ and fine-tune to multiple timesteps \cite{9950361,yao_attention_Pami}. With respect to scaling down timesteps during inference, one idea is to fine-tune a trained multi-timestep model to a single timestep \cite{chowdhury2021one}, or employ additional controllers to adjust inference timesteps adaptively \cite{yao2021temporal,li2023seenn,ding2024shrinking}.

\begin{figure*}[t]
% \vskip 0.2in
\begin{center}
\centerline{\includegraphics[width=\linewidth]{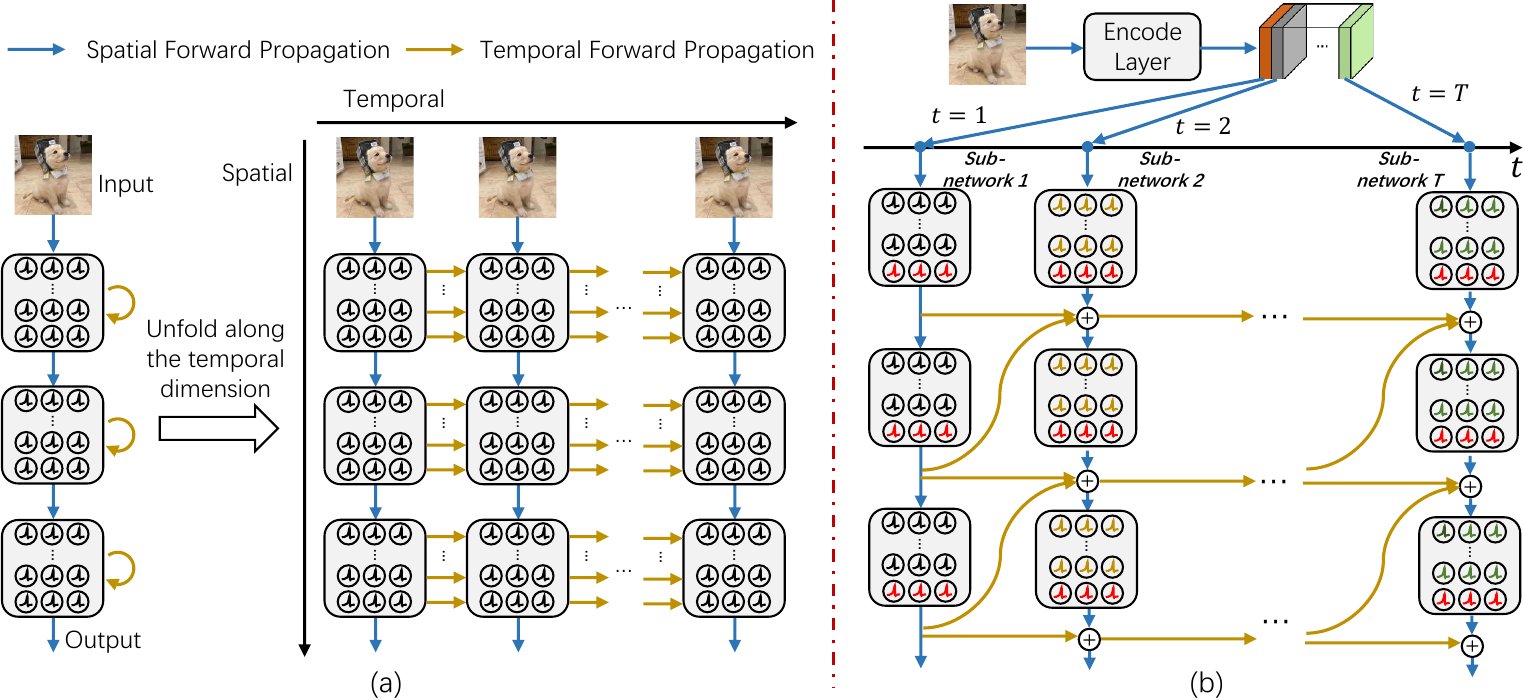}}
\vspace{-3mm}
\caption{Illustration of the temporal forward of vanilla SNN and T-RevSNN. (a) Vanilla SNNs unfold along the temporal, reusing all parameters at each timestep. All spiking neurons accomplish the temporal dynamics. In the image classification task, images are input repeatedly. Thus, the memory and inference costs of vanilla SNNs are $\mathcal{O}(L \times T)$ and $\mathcal{O}(T)$, respectively. (b) In T-RevSNN, we only allow the key spiking neurons (red spiking neurons in the figure) to pass temporal information. Coupled with the multi-level temporal reversible design, the memory cost of T-RevSNN is $\mathcal{O}(L)$. Moreover, the image is encoded only once. The encoded features are divided into $T$ groups, exploited as input for each timestep. Correspondingly, the entire SNN is also divided into $T$ independent sub-networks, which only share parameters and transfer temporal information at key spiking neurons. Thus, the inference of T-RevSNN is $\mathcal{O}(1)$.}
\label{Fig_1_overview}
\end{center}
\vspace{-1cm}
\end{figure*}

However, none of them can simultaneously achieve cheap training memory and low inference energy because they tend to optimize only in one direction. In this work, we can shoot two birds with one stone via a novel temporal reversible architecture for SNNs. Our motivation is simple and intuitive. It has been shown that the BPTT of SNNs through the temporal contributes just a little to the final gradients. \citet{xiao2022online} and \citet{meng2023towards} therefore do not calculate temporal gradients to improve training speed. That being the case, \emph{can we only retain the temporal forward at key positions and turn off the temporal dynamics of other spiking neurons?}

Based on this simple idea, we design temporal reversible SNNs (Fig.~\ref{Fig_1_overview}(b)). \emph{First}, to reduce training memory, we only turn on the temporal dynamics of the output spiking layer of each stage and design the temporal transfer to be reversible. Specifically, reversibility \cite{gomez2017reversible} means that we do not need to store the membrane potentials and activations of all spiking neurons to compute gradients. \emph{Second}, since turned-off spiking neurons have no temporal dynamics, we simply let these neurons no longer reuse parameters in the temporal dimension. In order not to increase parameters and energy cost, we encode the input only once and divide the input features and the entire network into $T$ groups (Fig.~\ref{Fig_1_overview}(b)). \emph{Third}, to improve performance, the information transfer between stages is carried out in a multi-level form. We also redesign the SNN blocks in the style of ConvNeXt \cite{liu2022convnet} and adjust the residual connection to ensure the effectiveness of information transfer. In summary, the backward propagation of SNNs will be altered by their forward, resulting in cheap memory and inference costs concurrently. 

We verify the effect of T-RevSNN on static (ImageNet-1k \cite{deng2009imagenet}) and neuromorphic datasets (CIFAR10-DVS \cite{li2017cifar10} and DVS128 Gesture \cite{amir2017dvsg}). Our contributions are:

% \vspace{-5mm}

\begin{itemize}
\item We redesign the forward propagation of SNNs simply and intuitively to simultaneously achieve low training memory, low power, and high performance.
% \vspace{-5mm}

\item We have made systematic designs in three aspects to implement the proposed idea, including multi-level temporal-reversible forward information transfer of key spiking neurons, group design of input encoding and network architecture, and improvements in SNN blocks and residual connections. 
% \vspace{-5mm}

\item On ImageNet-1k, our model achieves state-of-the-art accuracy on CNN-based SNNs with minimal memory and inference cost and the fastest training. Compared to current Transformer-based SNN with state-of-the-art performance, i.e., spike-driven Transformer \cite{yao2023spikedriven}, our model achieves close accuracy, while the memory efficiency, training time acceleration, and inference energy efficiency can be significantly improved by $8.6 \times$, $2.0 \times$ and $1.6 \times$, respectively.

\end{itemize}

\section{Related Work}

\textbf{Training of SNNs.} The two primary deep learning techniques for SNNs are ANN-to-SNN conversion (ANN2SNN) \cite{wu2021progressive,9597475} and direct training \cite{wu2018spatio,neftci2019surrogate}. The ANN2SNN method approximates the activations in ANNs using firing rates in SNNs. It can achieve accuracy close to ANNs, but correct firing rate estimation takes a lot of timesteps \cite{sengupta2019going,Rathi2020Enabling,hu2023fast,wu2021progressive,wang2023masked,guo2023direct}. The direct training approach uses the surrogate gradient to solve the non-differentiable problem caused by binary spike firing. Through the BPTT, SNNs' training only requires a few timesteps. There is a performance gap between directly training SNNs and ANNs, which has been shrinking \cite{fang2021deep,hu2024advancing,yao_attention_Pami,zhou2023spikformer,yao2023spikedriven,qiu2023vtsnn}. In this work, we focus on direct training.

\textbf{Improvement of Architecture and Training in SNNs.} SNNs have become larger and deeper. This is due to improvements in architecture and training. Spiking ResNet \cite{hu2024advancing,fang2021deep} and Transformer \cite{zhou2023spikformer,yao2023spikedriven} are two benchmark architectures, inspired by classic ResNet \cite{he_resnet_2016,He_2016_identitymapping} and Transformer \cite{transformer,dosovitskiy2021an}, respectively. Various training methods have been proposed to improve accuracy, such as normalization \cite{zheng2021going,kim2021revisiting,duan2022temporal}, gradient optimization~\cite{perez2021sparse,ijcai2023p335}, attention \cite{yao2021temporal,qiu2024gated,DENG2024111780,yao2023inherent,yao2023sparser}, membrane optimization \cite{guo2022imloss,guo2022reducing,guo2022recdis}. The memory required to train large-scale SNNs is much higher than that of ANNs because of the storage of activation values over multiple timesteps. Some works are devoted to reducing the memory cost for training, including Online Training Through Time (\textbf{OTTT}) \cite{xiao2022online}, Spatial Learning Through Time (\textbf{SLTT}) \cite{meng2023towards}, etc. We compare our T-RevSNN with these training optimization methods in Section~\ref{Chap_complex_compare}.

% , Table~\ref{tab_complexity_analysis} and Table~\ref{Table_imagenet}.

\textbf{Reversible Neural Networks.} The goal of reversible neural networks \cite{gomez2017reversible} is to reduce memory cost when training the network. The core idea is that the activations of each layer can be calculated based on the subsequent reversible layer's activations. This enables a reversible neural network to perform backpropagation without storing the activations in memory, with the exception of a few non-reversible layers. The reversible idea has been applied to various architectures, such as Recurrent Neural Networks (RNNs) \cite{mackay2018reversible}, ResNet \cite{sander2021momentum,cai2023reversible}, vision transformer \cite{mangalam2022reversible}, U-Net \cite{brugger2019partially}, etc. Recently, the idea of reversibility has been applied to SNN \cite{RevSNN}, which is spatially reversible and we named it \textbf{S-RevSNN}. In contrast, our method is temporal-reversible.

\section{Preliminaries}\label{Chap_pre}

\textbf{Spiking neurons} have neuronal dynamics that enable spatio-temporal information processing. The dynamics of spiking neurons can be characterized~\cite{xiao2022online}:

% \begin{equation}
\begin{align}
\label{eq:linear_trans} V^l[t+1] &= (1 - \frac{1}{\tau_m}) V^l[t] + \frac{1}{\tau_m} \mathbf{W}^l S^{l-1}[t+1], \\
\label{eq:spike} S^l[t+1] &= \Theta(V^l[t+1] - V_{th}),
\end{align}
% \end{equation}

where $l = 1, \cdots, L$ is the layer index, $t$ is the timestep, $S^{l-1}[t]$ and $V^l[t-1]$ are the spatial input spikes and the temporal membrane potential input respectively, $\mathbf{W}^l$ is synaptic weight, $\tau_m$ is the decay factor, $V_{th}$ is the threshold. The Heaviside step function is denoted by $\Theta(x)$. For every $x \geq 0$, $\Theta(x)$ = 1; otherwise, $\Theta(x) = 0$. The input spatio-temporal information is incorporated into $V^l[t]$ and then compared with the threshold to decide whether to fire spikes.

\textbf{Spatio-Temporal BackPropagation (STBP)} is a combination of BPTT and surrogate gradient \cite{wu2018spatio,neftci2019surrogate}, where the latter is used to solve the non-differentiable of spike signals. Specifically, the gradients for weight $\mathbf{W}^l$ at timestep $T$ is calculated by
\begin{equation}\label{eq:stbp}
\nabla_{\mathbf{W}^l} \mathcal{L} = \sum_{t=1}^{T} (\frac{\partial \mathcal{L}}{\partial V^l[t]}) ^\top S^{l-1}[t]^\top
\end{equation}
where $\mathcal{L}$ is the loss. It can be calculated by:
\begin{align}
\label{eq:unfold_stbp}
\frac{\partial \mathcal{L}}{\partial  V^{l}[t]}&=
\frac{\partial \mathcal{L}}{\partial S^{l}[t]}
{\frac{\partial  S^{l}[t]}{\partial  V^{l}[t]}}+
\frac{\partial \mathcal{L}}{\partial  V^{l}[t+1]} 
{\frac{\partial  V^{l}[t+1]}{\partial  V^{l}[t]}},
\end{align}
where
\begin{align}
\label{eq:unfold_stbp_2}
     \frac{\partial \mathcal{L}}{\partial  S^{l}[t]}&=
     \frac{\partial \mathcal{L}}{\partial V^{l+1}[t]} \frac{\partial V^{l+1}[t]}{\partial  S^{l}[t]}
     +\frac{\partial \mathcal{L}}{\partial V^{l}[t+1]} {\frac{\partial V^{l}[t+1]}{\partial S^{l}[t]}}.
\end{align}
The non-differentiable term ${\frac{\partial S^{l}[t]}{\partial V^{l}[t]}}$ is the spatial gradient. It can be replaced by the surrogate function. ${\frac{\partial V^{l}[t+1]}{\partial S^{l}[t]}}$  and ${\frac{\partial  V^{l}[t+1]}{\partial  V^{l}[t]}}$ are the temporal gradients that need to be calculated. According to Eq.~\ref{eq:stbp} and Eq.~\ref{eq:unfold_stbp}, exploiting the STBP algorithm requires storing the activations and membrane potentials of all spiking neurons at all timesteps.

\section{Methods}\label{Chap_method}
\begin{figure*}[!t]
\centering
\includegraphics[width=0.99\linewidth]{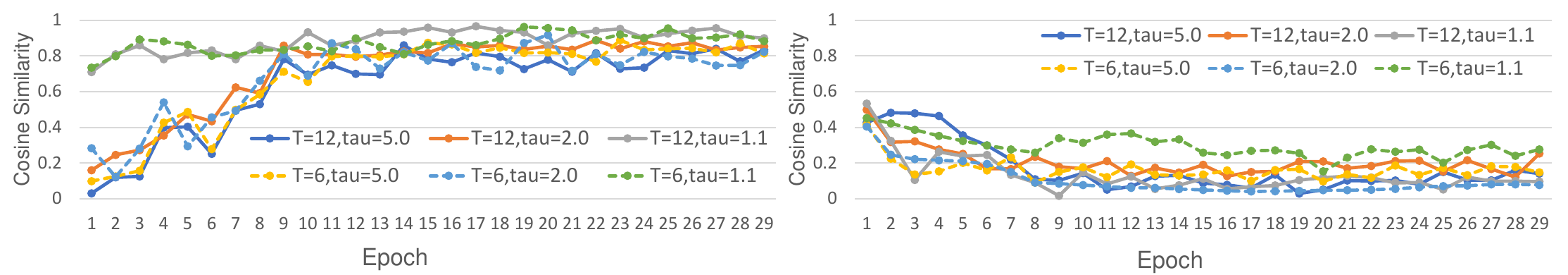}
\vspace{-2mm}
\caption{Left and Right sub-figures are the cosine similarity between the spatial gradients calculated by baseline and case 1/2, respectively. In case 1/2, we only retain/discard the temporal gradients of the last layer of spiking neurons in each stage. We use spiking ResNet-18 to train on CIFAR-10. $T$ and $\tau$ are Timestep and decay, respectively. The temporal gradients of the final layer of each stage are more significant (Case 1, left sub-figure, high cosine similarity) than those of spiking neurons in preceding stages (Case 2, right sub-figure, low cosine similarity). Due to space constraints, cosine similarity calculation details are given in the supplementary material.}
\label{fig:cos1}
\vspace{-5mm}
\end{figure*}
\subsection{Motivation}
Previous works \cite{xiao2022online, meng2023towards} observed that the spatial gradients dominate the final calculation of derivatives. Thus, they abandon the calculation of temporal gradients to speed up training. In contrast, our idea is to turn off the forward propagation of most spiking neurons in time, so that the computation of the temporal gradient is naturally drastically reduced. Building upon these insights, we analyze which spiking neurons are relatively important for temporal information transfer.

We set up the following experiments. We first use STBP to train SNN on CIFAR-10. The calculated gradient is called the ``baseline gradient", which needs to maintain the spatial and temporal computational graph of the previous timestep. In general, a typical neural network is divided into four stages, and the feature levels of each stage are various \cite{he_resnet_2016}. An intuitive idea is that the transfer of temporal information at the junction of two stages in SNNs may be important. Thus, we train two other SNNs using STBP. One (\emph{Case 1}) discards most of the temporal gradients during training, retaining only the temporal gradients of the last layer in each stage. The other (\emph{Case 2}) adopts the opposite behavior, discarding only the temporal gradient of the last spiking neuron layer of each stage.

Then, we compare the difference between the baseline, case 1, and case 2 by calculating the cosine similarity of the spatial gradient across the network. Results are shown in Fig.~\ref{fig:cos1}. The similarity in the left sub-figure (comparison between baseline and case 1) has always maintained a high level, especially as the training epochs increase. In contrast, The similarity in the right sub-figure (comparison between baseline and case 2) is always small. Thus, the temporal gradients of the final layer of each stage are more significant than those of spiking neurons in preceding stages. 

\begin{figure}[h]
\begin{center}
\centerline{\includegraphics[width=0.9\columnwidth]{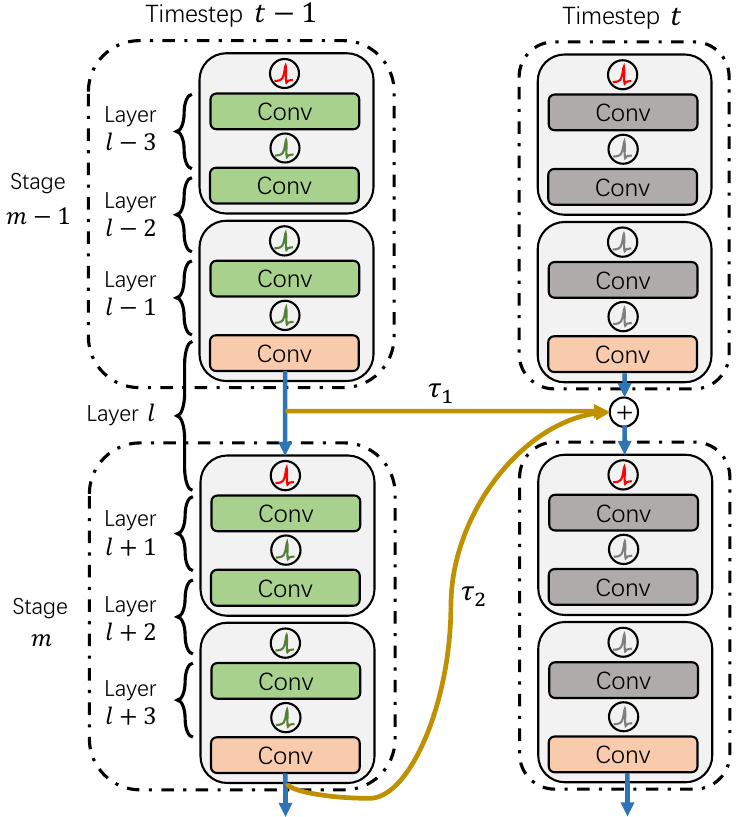}}
\caption{Temporal-reversible connection in T-RevSNN.}
\label{Fig_TRevSNN_Forward}
\end{center}
\vskip -0.3in
\end{figure}

\subsection{Temporal-Reversible Spike Neuron}
\label{4.2}
Based on the above analysis, we only turn on the temporal transfer of the spiking neurons in the last layer of each stage, as shown in Fig.~\ref{Fig_TRevSNN_Forward}. Thus spiking neurons in T-RevSNN have two states: turn-on or turn-off in the temporal dimension. The forward of temporal turn-off spiking neurons is
\begin{equation}\label{eq:off_forward}
V^l[t] = \mathbf{W}^l S^{l-1}[t],
\end{equation}
and the weight update depends solely on the spatial gradient: 
\begin{equation}\label{eq:unimportant_neuron_backward}
\nabla_{\mathbf{W}^l} \mathcal{L} = \sum_{t=1}^{T}
(\frac{\partial \mathcal{L}}{\partial V^{l+1}[t]} \frac{\partial V^{l+1}[t]}{\partial  S^{l}[t]} 
\frac{\partial  S^{l}[t]}{\partial  V^{l}[t]})^{\top} S^{l-1}[t]^{\top}
\end{equation}
When updating turn-off spiking neuron weights $\mathbf{W}^l$, the gradient needs to be calculated on the sub-network at each timestep. We disable weight reuse of turn-off spiking neurons in the temporal dimension. Therefore, the gradients for weight $\mathbf{W}^l[t]$ at layer $l$, timestep $t$ can be simplified by
\begin{equation}\label{eq:yunimportant_neuron_backward_single}
\nabla_{\mathbf{W}^l[t]} \mathcal{L} =
(\frac{\partial \mathcal{L}}{\partial V^{l+1}[t]} \frac{\partial V^{l+1}[t]}{\partial  S^{l}[t]} 
\frac{\partial  S^{l}[t]}{\partial  V^{l}[t]})^{\top} S^{l-1}[t]^{\top},
\end{equation}
which means that the update of all weights at timestep $t$ only needs to perform $\mathcal{O}(L)$ calculation.

For temporal turn-on spiking neurons, the forward function is Eq.~\ref{eq:linear_trans} and the weight update depends on both spatial and temporal gradients (i.e., Eq.~\ref{eq:unfold_stbp}). Inspired by the concept of reversibility \cite{gomez2017reversible}, we observe that Eq.~\ref{eq:linear_trans} is naturally reversible. Eq.~\ref{eq:linear_trans} can be rewritten as: 
\begin{equation}\label{eq:inverse_time}
% \begin{aligned}
% \text{Forward: } & V^l[t] = (1 - \frac{1}{\tau_m}) V^l[t-1] + \frac{1}{\tau_m} \mathbf{W}^l S^{l-1}[t] \\ 
V^l[t] = (1 - \frac{1}{\tau_m})^{-1} ( V^l[t+1] - \frac{1}{\tau_m} \mathbf{W}^l S^{l-1}[t+1]).
% \end{aligned}
\end{equation}
Subsequently, a reversible transformation can be established between $V^l[t+1]$ and $V^l[t]$. This means that when calculating the gradient at the first timestep, there is no need to store the membrane potentials and activations at all timesteps. We only need to store $V^l[T]$, and we can reversely deduce the membrane potential and activation values of all previous timesteps through Eq.~\ref{eq:inverse_time}. This reduces the memory required for multi-timestep training of SNNs. Regarding to the time complexity, the gradient process of the temporal turn-on spiking neurons is consistent with STBP, i.e., $\mathcal{O}(T^2)$. 

Multi-level feature interaction has been proven to improve the model effectiveness in architectural design \cite{Lin_2017_CVPR,gao2019res2net}. The spatio-temporal neuronal dynamics of SNNs already naturally contain two levels of features, where spatial inputs come from low-level features  (the previous layer at the same timestep)  and temporal inputs come from same-level features (the previous timestep in the same layer). And, existing residual connections in SNNs \cite{fang2021deep,hu2024advancing} also bring low-level features. But there's more we can do. In this work, we establish stronger multi-level connections between SNNs at adjacent timesteps. We incorporate the high-level features in deeper layers of the previous timestep into the information fusion of the current timestep. Generally, we can construct forward information passing as follows:
\begin{equation}
\label{Eq_multi_feature}
V^l[t+1] = \sum^{L}_{i=l} (1 - \frac{1}{\tau_{i}}) V^i[t] + \frac{1}{\tau_m} \mathbf{W}^l S^{l-1}[t+1], \\
\end{equation}
where $\tau_i$ means the decay factor at different layers and $\tau_l  = \tau_m$. Eq.~\ref{Eq_multi_feature} maintains temporal reversibility as follows:
\begin{equation}
\begin{aligned}
V^l[t] = (1 - \frac{1}{\tau_m})^{-1} ( \sum^{L}_{i=l+1}(1 - \frac{1}{\tau_i}) (V^i & [t+1] \\
 - \frac{1}{\tau_m} \mathbf{W}^l S^{l-1} & [t+1]) ). \\
\end{aligned}
\end{equation}
Note, as shown in Fig.~\ref{Fig_TRevSNN_Forward}, we only build high-level feature incorporation between stages. In multi-level interaction, we use up/down sampling to achieve feature alignment. 

\subsection{Basic SNN Block}
\begin{figure}[t]
\begin{center}
\centerline{\includegraphics[width=0.4\columnwidth]{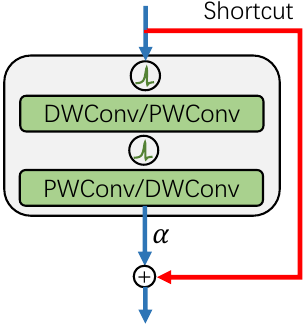}}
\caption{Basic SNN Block, following ConvNeXt-style \cite{liu2022convnet}.}
\label{Fig_basic_SNN_block}
\end{center}
\vskip -0.3in
\end{figure}

As shown in Fig.~\ref{Fig_basic_SNN_block}, we design the basic SNN Block following~\cite{liu2022convnet} to extract local information effectively. It consists of two depth-Wise separable convolutions (DWConv/PWConv) and a residual connection, where the kernel size of DWConv is $5 \times 5$. We drop all Batch Normalization (BN) \cite{ioffe2015batch} modules to reduce the memory cost caused by storing the mean and variance of features~\cite{xiao2022online}. To address the instability of training caused by loss of BN, the weights of all layers in the network are properly normalized~\cite{brock2021high}. In addition, the ReZero~\cite{bachlechner2021rezero} technique is used on the Membrane Shortcut \footnote{In addition to Membrane Shortcut \cite{hu2024advancing}, another type of residual commonly used in SNNs is Spike-Element-Wise (SEW) \cite{fang2021deep}, which cannot utilize this technique due to its nonlinear operations and inability to be re-parameterization.} \cite{hu2024advancing} to enhance the network's capacity to meet dynamic isometrics after initialization and to facilitate efficient network training. To guarantee that only addition operations occur in inference, we can merge the scale of ReZero (i.e., $\alpha$ in Fig.~\ref{Fig_basic_SNN_block}) into the weight $\mathbf{W}^l$ at layer $l$ through re-parameterization. 

\begin{figure*}[t]
\vskip -0.2in
\begin{center}
\centerline{\includegraphics[width=\linewidth]{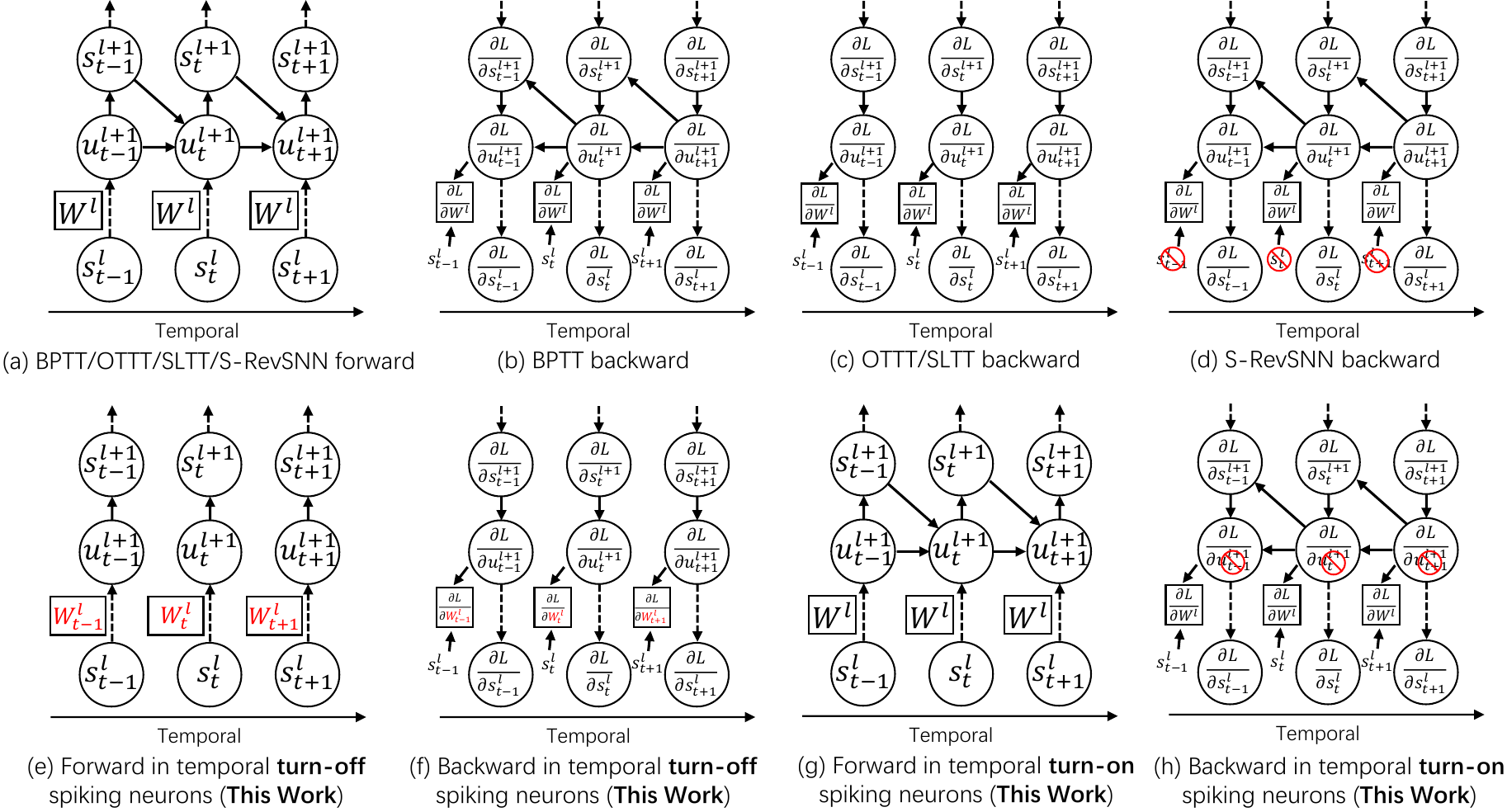}}
\vspace{-3mm}
\caption{Illustration of the forward and backward. (a) Existing training methods do not change the forward of SNNs. (b) The backward of STBP unfolds simultaneously along the temporal and spatial dimensions. (c) The backward of OTTT/SLTT only unfolds along the spatial dimension. (d) The backward of S-RevSNN unfolds along the temporal and spatial dimensions but is reversible in spatial. (e) and (f) show the forward and backward of the temporal turn-off spiking neurons in the proposed method, respectively. (g) and (h) give the forward and backward of the temporal turn-on spiking neurons in T-RevSNN, which are basically consistent with the forward and backward in (a) and (b). The difference is that the backward in (h) is reversible, so only the membrane potentials of the last timestep needs to be stored.}
\label{Fig_Forward_Backward}
\vspace{-1cm}
\end{center}
\end{figure*}

\subsection{Training and Inference Complexity Analysis}\label{Chap_complex_compare}

The memory and computation required by the STBP algorithm to calculate the gradient from \emph{the output of the last layer at the last timestep} to \emph{the input of the first layer at the first timestep} constitute the memory and time complexity of training SNNs. We analyze the training memory and time complexity of the proposed T-RevSNN and other SNN training optimization methods in Table~\ref{tab_complexity_analysis} and Fig.~\ref{Fig_Forward_Backward}.

% 我们首先定义，

% 在推理时

\begin{table}[H]
\centering
\scalebox{0.88}{
\begin{tabular}{c | c | c | c}
\hline
\multirow{2}*{Training Methods} & \multicolumn{2}{c|}{Training} & Inference \\ \cline{2-4}
~ & Memory & Time & Energy \\ \hline
% ~ & Cost & Complexity & ~ \\ \hline
STBP\small{~\cite{neftci2019surrogate}} & $\mathcal{O}(LT)$ & $\mathcal{O}(LT)$ & $\mathcal{O}(T)$ \\
OTTT\small{~\cite{xiao2022online}} & $\mathcal{O}(L)$ & $\mathcal{O}(LT)$ & $\mathcal{O}(T)$ \\
SLTT-K\small{~\cite{meng2023towards}} & $\mathcal{O}(L)$ & $\mathcal{O}(LK)$ & $\mathcal{O}(T)$ \\
S-RevSNN & \multirow{2}*{$\mathcal{O}(T)$} & \multirow{2}*{$\mathcal{O}(T)$} & \multirow{2}*{$\mathcal{O}(T)$} \\
\small{~\cite{RevSNN}} & ~ & ~ & ~ \\ \hline
\textbf{T-RevSNN turn-off (Ours)} & \multirow{2}*{$\mathcal{O}(L)$} & $\mathcal{O}(L)$ & \multirow{2}*{$\mathcal{O}(1)$} \\
\textbf{T-RevSNN turn-on (Ours)} & ~ & $\mathcal{O}(T)$ & ~ \\ \hline
\end{tabular}
}
\caption{Training and inference complexity analysis.}
% \vskip -0.3in
\label{tab_complexity_analysis}
\end{table}

\textbf{STBP} \cite{wu2018spatio,neftci2019surrogate} needs to retain the activations of all spiking neurons at all timesteps, the training memory is $\mathcal{O}(LT)$. The temporal gradient is propagated between the two layers of spiking neurons. When updating the weights of layer $l$, $T$ gradients need to be calculated and summed up. It implies that the time complexity is $\mathcal{O}(T)$ per layer, so the overall training time complexity of STBP is $\mathcal{O}(LT)$. Please see Fig.~\ref{Fig_Forward_Backward} (b).

\textbf{OTTT} \cite{xiao2022online} does not calculate the temporal gradient and can be trained online. Each timestep only needs to retain activations at the current timestep, so the training memory and time are $\mathcal{O}(L)$ and $\mathcal{O}(LT)$, respectively.

\textbf{SLTT-K} \cite{meng2023towards} only calculates the gradients on the most important $K (K <= T)$ timesteps based on OTTT, thus the training memory and time complexity are $\mathcal{O}(L)$ and $\mathcal{O}(LK)$, respectively. Please see Fig.~\ref{Fig_Forward_Backward} (c).

\textbf{S-RevSNN} \cite{RevSNN} retains the temporal gradient. Due to spatial reversibility, S-RevSNN only needs to store the activations of the spatially reversible layer at each timestep. Thus, the training memory and time complexity are both $\mathcal{O}(T)$. Please see Fig.~\ref{Fig_Forward_Backward} (d).

\textbf{T-RevSNN (This work).} We discard most of the temporal connections and are temporally reversible, so training memory of T-RevSNN is consistent with OTTT/SLTT, i.e., $\mathcal{O}(L)$. The training time of T-RevSNN is related to spiking neurons (turn-on/off), see Fig.~\ref{Fig_Forward_Backward} (e) to (h).

% T-RevSNN turn-on neurons have temporal dynamics, which makes its performance in backpropagation completely consistent with that of normal SNN neurons trained with STBP. Since there is only one layer containing the turn-on neurons in each stage, the time complexity of the T-RevSNN turn-on is $O(T)$, see Fig.~\ref{Fig_Forward_Backward} (e) to (h).

The optimization of OTTT/SLTT/S-RevSNN does not change the forward propagation, so their inference is all $\mathcal{O}(T)$. The proposed T-RevSNN redign the input encoding and split the network, thus the inference is $\mathcal{O}(1)$.

\section{Experiments}
We evaluate the proposed T-RevSNN on static ImageNet-1k~\cite{deng2009imagenet} and neuromorphic CIFAR10-DVS~\cite{li2017cifar10}, DVS128 Gesture~\cite{amir2017dvsg}. We generally continued the experimental setup in~\cite{yao2023spikedriven}. The input size of ImageNet is $224 \times 224$. The batch size is set to 512 during 300 training epochs with a cosine-decay learning rate whose initial value is 0.002. The optimizer is Lamb. Standard data augmentation techniques, like random augmentation, are also employed in training. Details of the datasets, training and experimental setup, and power analysis are given in the supplementary material. 

\subsection{Main Results}

% Results of ImageNet-1k and CIFAR10-DVS/Gesture are given in Table~\ref{Table_imagenet} and Table~\ref{dvsdataset}, respectively.

\subsubsection{ImageNet-1k}

\begin{table*}[tp]
\centering
\scalebox{0.9}{
\begin{tabular}{lccccccc}
\hline
\multirow{2}{*}{Methods} & \multirow{2}{*}{Architecture} & Param & Time & Training Time & Memory & Energy & \multirow{2}{*}{Acc (\%)} \\ 
~ & ~ & (M) & step & (min/ep) & (MB/img) & (mJ) & ~ \\ \hline
% Hybrid training~\cite{Rathi2020Enabling} & ResNet-34 & 21.8 & 250 & - & - & - & 61.5 \\
% Diet-SNN~\cite{9556508} & VGG-16 & 38.1 & 5 & - & - & - & 69.0 \\ \hline
STBP-tdBN~\cite{zheng2021going} & ResNet-34 & 21.8 & 6 & 29.6 & 186.1 & 6.4 & 63.7 \\ \cline{2-8}
% \multirow{2}*{SEW ResNet~\cite{fang2021deep}} & SEW-ResNet-34 & 21.8 & 4 & 5.0 & 224.5 & 4.0 & 67.0 \\ 
SEW ResNet~\cite{fang2021deep} & SEW-ResNet-50 & 25.6 & 4 & 10.0 & 596.9 & 4.9 &  67.8 \\ \cline{2-8}
% \multirow{2}*{MS ResNet~\cite{hu2024advancing}} & MS-ResNet-18 & 11.7 & 6 & 4.6 & 185.9 & 4.3 & 63.1 \ \
MS ResNet~\cite{hu2024advancing} & MS-ResNet-34 & 21.8 & 6 & 11.2 & 267.1 & 5.1 & 69.4 \\ \cline{2-8}
TEBN~\cite{duan2022temporal} & ResNet-34 & 21.8 & 4 & 16.3 &  260.1 & 6.4* & 64.3  \\ \cline{2-8}
TET \cite{deng2022temporal} & SEW-ResNet-34 & 21.8 & 4 & 12.5 & 221.0 & 4.0* & 68.0 \\ \hline
Spikformer & Spikeformer-8-384 & 16.8 & 4 & 14.2 & 580.8 & 7.7 & 70.2 \\ 
\cite{zhou2023spikformer} & Spikeformer-8-512 &  29.7 & 4 &  16.7 & 767.8 & 11.6 & 73.4 \\ \cline{2-8}
Spike-driven Transformer & Spikeformer-8-384 & 16.8 & 4 & 15.4 & 548.9 & 3.9 & 72.3 \\ 
\cite{yao2023spikedriven} & Spikeformer-8-512 & \textbf{29.7} & 4 &  18.8 & 730.0 & 4.5 & \textbf{74.6} \\ \hline

OTTT~\cite{xiao2022online} & ResNet-34 & 21.8 & 6 & 24.2 & 84.1 & 6.0* & 64.2 \\ \cline{2-8}
\multirow{2}*{SLTT~\cite{meng2023towards}} & ResNet-34 & 21.8 & 6 & 18.1 & 71.7 & 6.0* & 66.2 \\
~ & ResNet-50 & 25.6 & 6 & 23.4 & 117.3 & 7.2* & 67.0 \\ \cline{2-8}
% SLTT~\cite{meng2023towards} & NF-ResNet-101 & - & 6 & - & - & - & 69.1 \\ \hline
% \multirow{2}*{Parallel SNN~\cite{fang2023parallel}} & SEW-ResNet-18 & 11.7 & 4 & 5.8 & 138.7 & - & 67.6 \\
% ~ & SEW-ResNet-34 & 21.8 & 4 & 8.3 & 179.7 & 4.0* & 70.5  \\ \cline{2-8}
\multirow{2}*{S-RevSNN~\cite{RevSNN}} & Rev-ResNet-37 & 24.4 & 4 & 7.9 & 127.1 & - & - \\ 
~ &  Rev-SFormer-8-384 & 17.2 & 4 & 16.3 & 389.6 & - & - \\ \hline

\multirow{2}*{\textbf{T-RevSNN (This work)}} & ResNet-18 (384) & 15.2 & 4 & \textbf{6.1} & \textbf{57.5} & \textbf{1.7} & 69.8 \\
~ & ResNet-18 (512) & \textbf{29.8} & 4 & 9.1 & 85.7 & 2.8 & \textbf{73.2} \\ \hline
% \multirow{2}*{\textbf{T-RevSNN-2 (This work)}} & NF-ResNet-18 (256) & 15.7 & 4 & 5.8 & \textbf{56.7} & \textbf{1.4} & 70.1 \\
% ~ & NF-ResNet-18 (384) & 30.2 & 4 & 8.2 & \textbf{83.1} & \textbf{2.8} & 73.1 \\ \hline
\end{tabular}
}
% \vspace{-1mm}
\caption{Evaluation on ImageNet-1K. We divide the existing direct training SNNs into three types, from top to bottom: spiking ResNet, spiking Transformer, and SNNs with training optimization (OTTT/SLTT/S-RevSNN). We measure the memory consumption of all models and the time it takes to complete an epoch. The memory per image (MB/img) is measured as the peak GPU memory each image occupies during training, following S-RevSNN~\cite{RevSNN}. All experiments were tested under float16 automatic mixed precision, distributed data-parallel, and no gradient checkpoint. * We estimate the power of these models based on the corresponding spiking ResNet, e.g., we assume that the power of one timestep of SEW-ResNet-50 is $4.0/4 = 1.0$mJ. In our model, 384/512 represents the channel number in the last stage, which is used to control the parameters.}
\label{Table_imagenet}
% \vspace{-2mm}
\end{table*}

We comprehensively compare the proposed T-RevSNN with existing work in the five aspects of parameters, training time, memory, inference energy, Top-1 accuracy. In Table~\ref{Table_imagenet}, we divide the existing direct training SNNs into three types, namely spiking ResNet, spiking Transformer, and SNNs with training optimization. We test various SNNs on 6 NVIDIA-A100-40GB devices and report their memory and training time, to demonstrate the superiority of T-RevSNN.

\textbf{Compared with Spiking ResNet Baselines.} There are two main types of spiking ResNet baselines in the SNN domain, SEW-ResNet \cite{fang2021deep} and MS-ResNet \cite{hu2024advancing}. The proposed T-RevSNN achieves the best accuracy among the spiking ResNet series, with the lowest training memory and time, and inference energy. For example, \textbf{T-RevSNN (This work)} vs. SEW-ResNet-50 vs. MS-ResNet-34: Param, \textbf{29.8M} vs. 25.6M vs. 21.8M; Acc, \textbf{73.2\%} vs. 67.8\% vs. 69.4\%; Memory, \textbf{85.7 (MB/img)} vs. 596.9 (MB/img) vs. 267.1 (MB/img); Traning-Time, \textbf{9.1 min/ep} vs. 10.0 min/ep vs. 11.2 min/ep; Inference, \textbf{2.8mJ} vs. 4.9mJ vs. 5.1 mJ. Thus, the proposed T-RevSNN shows a full range of advantages over existing spiking ResNets.

\textbf{Compared with Spiking Transformer Baselines.} Spiking Transformer is a type of architecture that has become popular in SNNs recently. Typical works include SpikFormer\cite{zhou2023spikformer} and Spike-driven Transformer \cite{yao2023spikedriven}. The results in Table~\ref{Table_imagenet} show that under similar parameters (30M), the accuracy of T-RevSNN (73.2\%) is the same as SpikFormer (73.4\%), but 1.4\% lower than Spike-driven Transformer (74.6\%). \emph{We think this may be a performance gap caused by the architecture (Transformer vs. CNN).} In this work, our SNN blocks are entirely composed of convolutional layers. In contrast, Spike-driven Transformer contains the spiking self-attention operator. We tried replacing the spiking CNN block with spiking Transformer in \citet{zhou2023spikformer} and \citet{yao2023spikedriven}, but the network would not converge. And, T-RevSNN has overwhelming advantages in training speed and memory.

% We have theoretically analyzed the differences between the proposed T-RevSNN and OTTT/SLTT/S-RevSNN in forward/backward/training and inference complexities in Table~\ref{tab_complexity_analysis} and Fig.~\ref{Fig_Forward_Backward} (Section~\ref{Chap_complex_compare}).

\textbf{Compared with Training Optimization SNNs.} As can be seen in Table~\ref{Table_imagenet}, our model is better than existing training optimization SNNs in accuracy, training speed and memory, power. S-RevSNN \cite{RevSNN} does not report accuracy on ImageNet, so there is no relevant data.

Overall, T-RevSNN shows significant advantages over existing SNNs regarding training speed, memory requirements, and inference power. T-RevSNN also performs competitively in performance. Although the accuracy is lower than Spike-driven Transformer, we argue that this is a gap caused by the architecture and can be solved in the future.

\begin{table*}[tp]
\centering
\scalebox{0.9}{
\begin{tabular}{lccccc}
\hline
\multirow{2}{*}{Methods} & \multicolumn{2}{c}{CIFAR10-DVS} & \multicolumn{3}{c}{DVS128 Gesture} \\ \cmidrule(lr){2-3} \cmidrule(lr){4-6}
~ & Timestep & Top-1 Acc & Timestep & Top-1 Acc & Memory \\ \hline
LIAF-Net \cite{9429228} & 10 & 70.4 & 60 & 97.6 & - \\
Rollout  \cite{kugele2020efficient} & 48 & 66.8 & 240 & 97.1 & - \\
tdBN \cite{zheng2021going} & 10 & 67.8 & 40 & 96.9 & - \\
PLIF \cite{Fang_2021_ICCV} & 20 & 74.8 & 20 & 97.6 & - \\
SEW ResNet \cite{fang2021deep} & 16 & 74.4 & 16 & 97.9 & 81.5\\
MS ResNet \cite{hu2024advancing} & 10 & 76.0 & 10 & 94.8 & 79.4 \\
Dspike \cite{li2021differentiable} & 10 & 75.4 & - & - & - \\
~\cite{she2021sequence} & - & - & 20 & 98.0 & - \\
DSR \cite{meng2022training} & 10 & 77.3 & - & - & - \\ \hline
Spikformer \cite{zhou2023spikformer} & 16 & \textbf{80.6} & 16 & 97.9 & 51.8  \\
Spike-Driven Transformer \cite{yao2023spikedriven} & 16 & 80.0 & 16 & \textbf{99.3} & 51.4 \\ \hline

OTTT~\cite{xiao2022online} & 10 & 76.6 & 20 & 96.9 & - \\ 
SLTT~\cite{meng2023towards} & 10 & 77.2 & 20 & 97.9 & - \\ 
S-RevSNN~\cite{RevSNN} & 10 & 75.5 & 10 & 94.4 & 25.0 \\ \hline
\multirow{2}*{\textbf{T-RevSNN (This work)}} & 10 & 77.4 & 10 & 94.4 & \textbf{10.3} \\
~ & 16 & \textbf{79.2} & 16 & \textbf{97.9} & 18.9 \\ \hline

\end{tabular}
}
\vspace{-3mm}
\caption{Evaluation on CIFAR10-DVS \cite{li2017cifar10} and Gesture \cite{amir2017dvsg}. Memory is evaluated on the Gesture.}
\label{table_dvsdataset}
\vspace{-5mm}
\end{table*}

% Note that Gesture and CIFAR10-DVS are neuromorphic action classification datasets, which need to convert the event stream into frame sequences before processing. DVS128 Gesture is a gesture recognition dataset. CIFAR10-DVS is a neuromorphic dataset converted from CIFAR-10 by shifting image samples to be captured by the DVS camera. We keep the experimental setup in~\cite{yao2023spikedriven}, including the optimizer, learning rate, etc.

\subsubsection{CIFAR10-DVS and Gesture}
Unlike static datasets that encode repeated inputs, neuromorphic datasets input different information at each timestep \cite{deng2020rethinking}. Therefore, in the neuromorphic task, the inference complexity of all SNN models is $\mathcal{O}(1)$. We present the details of these two datasets as well as the pre-processing and experimental setup in the supplementary material. We still use the network architecture in Fig.~\ref{Fig_1_overview}(b). Different from encoding only once in static image tasks, here we encode the input at each timestep, which is consistent with prior SNNs. Experiments on CIFAR10-DVS and Gesture are shown in Table~\ref{table_dvsdataset}, we achieve performance comparable to SOTA results with lower memory.

\subsection{Ablation Studies}

We conduct various ablation studies on ImageNet to analyze the proposed T-RevSNN. Generally, we set the training epoch to 100 and keep the parameter to 15.2M.

\textbf{Timestep.} In our design, we divide the parameters of the entire network into $T$ groups (sub-networks). In Table~\ref{tab_different_timestep}, we analyze the impact of different $T$ on accuracy, training speed, and memory. Since we fix the total number of parameters, increasing the $T$ means that the sub-network at each timestep becomes smaller. The training memory decreases accordingly (training memory complexity of T-RevSNN is $\mathcal{O}(L)$), but the training time will increase. And, the relationship between accuracy and timestep is not linear.

% 同等参数下，不同时间步的能耗，精度，memory，速度。

% 由于在参数相同的情况下，时间步越大，每一时间步上的模型容量越小，O(L)下降，网络训练所需的memory将降低。

\vspace{-3mm}
\begin{table}[H]
\centering
\scalebox{0.85}{
\begin{tabular}{c|c|c|c}
\hline
\multirow{2}*{Timestep} & \multirow{2}*{Acc} & Memory & Training Time \\
~ & ~ & (MB/img) & (min/ep) \\ \hline
2 & 68.1 & 75.6 & \textbf{4.9} \\
4 & 68.6 & 57.5 & 6.1 \\
8 & \textbf{69.8} & 49.8 & 9.9 \\
16 & 63.4 & \textbf{42.4} & 16.0 \\
\hline
\end{tabular}
}
\vspace{-3mm}
\caption{Ablation study with timestep $T$.}
\label{tab_different_timestep}
\vspace{-5mm}
\end{table}

\textbf{Multi-level Temporal Reversible.} In Eq.~\ref{Eq_multi_feature}, we design multi-level feature fusion for T-RevSNN. We fuse the features of the next stage into the current stage. Otherwise, T-RevSNN's accuracy will lose 1.2\% (68.6\% $\rightarrow$ 67.4\%).

% \scalebox{0.85}{
% \begin{minipage}[c]{0.2\textwidth}
% \begin{tabular}{c|c}
% \hline
% Mutil-level & Acc \\ \hline
% \ding{51} & 68.6 \\
% \ding{55} & - \\ \hline
% \end{tabular}
% \captionof{table}{Ablation study with multi-level temporal reversible.}
% \end{minipage}
% \begin{minipage}[c]{0.5\textwidth}
% \begin{tabular}{c|c|c}
% \hline
% Scaled Residual & Acc & Epochs to 60\% Acc \\ \hline
% \ding{51} & 68.6 & 25 \\
% \ding{55} & 68.3 & 32 \\ \hline
% \end{tabular}
% \captionof{table}{Ablation study with residual connection.}
% \end{minipage}
% }

\textbf{Scaled Residual Connection} contributes to the convergence speed and final accuracy of the model, as shown in Table~\ref{tab_different_residual_connections}. 

\vspace{-3mm}
\begin{table}[H]
\centering
\scalebox{0.85}{
\begin{tabular}{c|c|c}
\hline
Scaled Residual & Acc & Epochs to 60\% Acc \\ \hline
\ding{51} & 68.6 & 25 \\
\ding{55} & 68.3 & 32 \\ \hline
\end{tabular}
}
\vspace{-3mm}
\caption{Ablation study with residual connection.}
\label{tab_different_residual_connections}
\vspace{-5mm}
\end{table}

\textbf{SNN Block Structure} is critical to final performance. In our design, we apply the idea of temporal reversibility to the ConvNeXt-style \cite{liu2022convnet} architecture. In Table~\ref{tab_different_structure}, we exploit the idea to MS-ResNet \cite{hu2024advancing} and S-RevSNN \cite{RevSNN}. When applying temporal reversibility directly on MS-ResNet-34, performance will drop a little. Moreover, we can see that temporal and spatial reversibility are orthogonal. The combination of the two can further improve training efficiency. Since S-RevSNN does not provide baseline accuracy on ImageNet, we only observe training efficiency here.

\vspace{-3mm}
\begin{table}[H]
\centering
\scalebox{0.85}{
\begin{tabular}{c|c|c|c}
\hline
\multirow{2}*{Block} & \multirow{2}*{Acc} & Memory & Training Time \\
~ & ~ & (MB/img) & (min/ep) \\ \hline
MS-ResNet-34 (Baseline) & 68.3 & 267.1 & 11.2 \\
+ T-RevSNN & 66.7 & 88.1 & 7.4 \\ \hline
S-RevSNN-37 (Baseline) & - & 127.1 & 7.9 \\
+ T-RevSNN & - & 55.4 & 6.8 \\ \hline
\end{tabular}
}
\vspace{-3mm}
\caption{Ablation study with SNN Block Structure.}
\label{tab_different_structure}
\vspace{-5mm}
\end{table}

\section{Conclusion}
We propose T-RevSNN to address the training memory and inference energy challenges posed by multiple timestep simulations for SNNs. Our idea is simple and intuitive. Since the temporal gradients in SNNs are not important, can we only keep the temporal forward direction of key positions and turn off the temporal dynamics of other spiking neurons? To this end, we make systematic designs, including multi-level temporal-reversible forward information transfer of key spiking neurons, group design of input encoding and network architecture, and improvements in SNN blocks and residual connections. Extensive experiments are conducted on static and neuromorphic datasets, verifying the advantages of T-RevSNN in training, inference and accuracy. We hope our investigations pave the way for further research on large-scale SNNs and more applications of SNNs.

% \clearpage

% Beijing Natural Science Foundation for Distinguished Young Scholars (JQ21015),

\textbf{Acknowledgement.} This work was partially supported by National Science Foundation for Distinguished Young Scholars (62325603), National Natural Science Foundation of China (62236009, U22A20103, 62441606), and CAAI-MindSpore Open Fund, developed on OpenI Community.

\textbf{Impact Statements.} This paper presents work whose goal is to advance the field of Machine Learning. There are many potential societal consequences of our work, none which we feel must be specifically highlighted here.

\bibliography{example_paper}
\bibliographystyle{icml2024}

%%%%%%%%%%%%%%%%%%%%%%%%%%%%%%%%%%%%%%%%%%%%%%%%%%%%%%%%%%%%%%%%%%%%%%%%%%%%%%%
%%%%%%%%%%%%%%%%%%%%%%%%%%%%%%%%%%%%%%%%%%%%%%%%%%%%%%%%%%%%%%%%%%%%%%%%%%%%%%%
% APPENDIX
%%%%%%%%%%%%%%%%%%%%%%%%%%%%%%%%%%%%%%%%%%%%%%%%%%%%%%%%%%%%%%%%%%%%%%%%%%%%%%%
%%%%%%%%%%%%%%%%%%%%%%%%%%%%%%%%%%%%%%%%%%%%%%%%%%%%%%%%%%%%%%%%%%%%%%%%%%%%%%%
\newpage
\appendix

\clearpage

\textbf{Limitation} of this work are Transformer-based T-RevSNN, larger scale models, more vision or long sequence tasks, and we will work on them in future work. The experimental results in this paper are reproducible. We explain the details of model training and configuration in the main text and supplement it in the appendix. Our codes and models will be available on GitHub after review. 
\par

\section{Details of  Cosine Similarity Calculation}
In Fig. \ref{fig:cos1}, the Left and Right sub-figures are the cosine similarity between the spatial gradients calculated by baseline and case 1/2, respectively.
In case 1/2, we only retain/discard the temporal gradients of the last layer of spiking neurons in each stage. We use spiking ResNet-18 to train on CIFAR-10 for 30 epochs with different hyperparameters (e.g., time step $T$, decay $\tau$). Moreover, we take the mean gradient value of each layer in every epoch during the training process as a vector and use the cosine similarity formula (Eq. \ref{cos}) to compare the differences between various training methods.
\begin{equation}
\label{cos}
    \mathcal{F}(A,B) = \frac{A \cdot B}{||A|| \ ||B||}=\frac{\sum\limits_{i=1}^{n}{A_i \cdot B_i}}{\sqrt{\sum\limits_{i=1}^{n}(A_i)^2} \cdot \sqrt{\sum\limits_{i=1}^{n}(B_i)^2}},
\end{equation}
where $\mathcal{F}(\cdot)$ is the cosine similarity between vector $A$ and $B$. $||A||$ and $||B||$ are the Euclidean norms of $A$ and $B$.

\section{Energy Consumption for T-RevSNN.}
The T-RevSNN architecture can transform matrix multiplication into sparse addition, which can be implemented as an addressable addition on neuromorphic chips. Moreover, when evaluating algorithms, the SNN field often ignores specific hardware implementation details and estimates theoretical energy consumption for a model \cite{panda2020toward,yin2021accurate,yang2022lead,KCP,meta_spikeformer}. This theoretical estimation is just to facilitate the qualitative energy analysis of various SNN and ANN algorithms. In the encoding layer, convolution operations serve as MAC operations that convert analog inputs into spikes, similar to direct coding-based SNNs \cite{wu2019direct}. Conversely, in SNN's architecture, the  Convolution (Conv) and Fully-Connected (FC) layers transmit spikes and perform AC operations to accumulate weights for postsynaptic neurons. The inference energy cost of T-RevSNN can be expressed as follows:
%\vspace{-2mm}
\begin{equation}
\label{energy}
    \begin{aligned}
E_{total}&=E_{MAC}\cdot FL_{conv}^1+\\&E_{AC}\cdot T \cdot(\sum_{n=2}^N FL_{conv}^n \cdot fr^{n} +\sum_{m=1}^M FL_{fc}^m \cdot fr^{m}),
\end{aligned}
\end{equation}
\par
where $N$ and $M$ are the total number of Conv and FC layers, $E_{MAC}$ and $E_{AC}$ are the energy costs of MAC and AC operations, and $fr^{m}$, $fr^{n}$, $FL_{conv}^n$ and $FL_{fc}^m$ are the firing rate and FLOPs of the $n$-th Conv and $m$-th FC layer. The spike firing rate is defined as the proportion of non-zero elements in the spike tensor. Previous SNN works  \cite{horowitz_energy_cost_2014, Rathi2020Enabling} assume 32-bit floating-point implementation in 45nm technology, where $E_{MAC}$ = 4.6pJ and $E_{AC}$ = 0.9pJ for various operations. 

\section{Experiment Details.} 

\textbf{Datasets.} We employ two types of datasets: static image classification and neuromorphic classification. The former includes ImageNet-1K \cite{deng2009imagenet}. The latter contains CIFAR10-DVS \cite{li2017cifar10} and DVS128 Gesture \cite{amir2017dvsg}.

ImageNet-1K is the most typical static image dataset, which is widely used in the field of image classification. It offers a large-scale natural image dataset of 1.28 million training images and 50k test images, with a total of 1,000 categories. CIFAR10-DVS is an event-based neuromorphic dataset converted from CIFAR10 by scanning each image with repeated closed-loop motion in front of a Dynamic Vision Sensor (DVS). There are a total of 10,000 samples in CIFAR10-DVS, with each sample lasting 300ms. The temporal and spatial resolutions are \textmu s and $128 \times 128$, respectively. DVS128 Gesture is an event-based gesture recognition dataset, which has the temporal resolution in \textmu s level and $128 \times 128$ spatial resolution. It records 1342 samples of 11 gestures, and each gesture has an average duration of 6 seconds.

\textbf{Data Preprocessing.} In the static datasets, as shown in Fig.~\ref{Fig_1_overview} (b), what differs from previous work is that we do not repeatedly feed the data into the SNN, but encode it and divide it into equal parts according to the number of timesteps. Then the features are each sent to the sub-networks. At the same time, due to the reversible connection of information between different timesteps, there is no information loss~\cite{gomez2017reversible,cai2023reversible} between these timesteps, so we do not need to average the output of all time steps before calculating the loss function. We only use the information from the last timestep to calculate the cross-entropy loss.

In contrast, neuromorphic (i.e. event-based) datasets can fully exploit the energy-efficient advantages of SNNs with spatio-temporal dynamics. Specifically, neuromorphic datasets are generated by event-based (neuromorphic) cameras, such as DVS~\cite{Gallego_2020_DVS_Survey,wu2022efficient,wu2022mss,xu2022transformers}. Compared to conventional cameras, DVS represents a new paradigm shift in visual information acquisition, encoding the time, location, and polarity of brightness changes for each pixel into event streams with \textmu s-level temporal resolution. Events (spike signals with address information) are generated only when the brightness of the visual scene changes. This is consistent with the event-driven nature of SNNs. Only when there is an event input will some spiking neurons of SNNs be triggered to participate in the computation. Typically, event streams are pre-processed into image sequences as input to SNNs. Details can be found in previous work~\cite{yao2021temporal}.

% \textbf{Encode layer and Classification heads.} In detail, the input is fed into a convolutional layer to extract the information. A LeakyReLU activation function is applied between these layers~\cite{chen2023vanillanet}, with its negative slope initialized to 0. During training, the negative slope will be adjusted according to the Eq.~\ref{eq:negative_slope}. After training, the negative slope will be fixed at 1. When the negative slope of LeakyReLU is 1, LeakyReLU degenerates into an identity mapping. At this point, due to the linear nature of convolution, the Encoder Layer can be re-parameterized into a single convolution, ensuring the networks can perform inference on neuromorphic chips. Similarly, we also use this technique on the classification layer following~\cite{chen2023vanillanet}. 

% \begin{equation}\label{eq:negative_slope}
% \text{negative slope} = 0.5 * (1 - cos(\frac{\pi * \text{epoch}}{100}))
% \end{equation}

\textbf{Experimental Steup.} The experimental setup in this work generally follows \cite{yao2023spikedriven}. The experimental settings of ImageNet-1K have been given in the main text. Here we mainly give the network settings on neuromorphic datasets. The training epoch for these datasets is 300. The batch size is 16 for Gesture and CIFAR10-DVS. The learning rate is initialized to 0.0003 for Gesture, and 0.01 for CIFAR10-DVS. All of them are reduced with cosine decay. We follow \cite{yao2023spikedriven} to apply data augmentation on Gesture and CIFAR10-DVS. In addition, the network structures used in CIFAR10-DVS and Gesture are T-RevSNN ResNet-18 (256).

\textbf{Detailed configurations and hyper-parameters.} We use the open-source timm~\cite{rw2019timm} repository for hyper-parameter configuration. We give here a detailed hyperparameter configuration to improve the reproducibility of T-RevSNN.

\begin{table}[h]
\vspace{-0.2cm}
\centering
% \vspace{3pt}
\begin{tabular}{c|c}
\hline
Hyper-parameter     & ImageNet    \\ \hline
Model size          & 15M/30M \\
\multirow{2}*{Embed dim}  & [64, 128, 256, 384] \\ 
~ & [96, 192, 384, 512] \\
Timestep            & 4           \\
Epochs              & 300          \\
Resolution          & 224*224     \\
Batch size          & 4096         \\
Optimizer           & LAMB        \\
Base Learning rate  & 2e-4    \\
Learning rate decay & Cosine      \\
Warmup eopchs       & 20           \\
Weight decay        & 0.01        \\
Rand Augment        & m9-mstd0.5-inc1       \\
Mixup               & None        \\
Cutmix              & None        \\
Label smoothing     & 0.1         \\ \hline
\end{tabular}
\caption{Hyper-parameters for image classification on ImageNet-1K.}
\label{table_train_imagenet_detail}
\end{table}

\onecolumn

\end{document}